\documentclass[runningheads]{llncs}

\usepackage{eccv}

\usepackage{eccvabbrv}

\usepackage{graphicx}
\usepackage{booktabs}

\usepackage[accsupp]{axessibility}  
\usepackage{multirow}
\usepackage{makecell}
\usepackage{xspace}
\usepackage{colortbl}
\usepackage[table]{xcolor}
\usepackage{arydshln}
\usepackage{subcaption}
\usepackage{algorithm}
\usepackage{algorithmic}
\usepackage{wrapfig}

\usepackage{hyperref}

\usepackage{orcidlink}

\begin{document}

\title{StepVAR: Structure-Texture Guided Pruning for Visual Autoregressive Models}
\def\methodNAME{StepVAR\xspace}

\titlerunning{StepVAR}

\newcommand*\samethanks[1][\value{footnote}]{\footnotemark[#1]}
\author{Keli Liu\thanks{Equal Contribution} \and Zhendong Wang\samethanks \and
Wengang Zhou \and
Houqiang Li}

\authorrunning{K.~Liu et al.}


\institute{
\vspace{-1em}
University of Science and Technology of China
\email{\{sa23006063,zhendongwang\}@mail.ustc.edu.cn}
\email{\{zhwg,lihq\}@ustc.edu.cn}}

\maketitle

\begin{abstract}
Visual AutoRegressive (VAR) models based on next-scale prediction enable efficient hierarchical generation, yet the inference cost grows quadratically at high resolutions. We observe that the computationally intensive later scales predominantly refine high-frequency textures and exhibit substantial spatial redundancy, in contrast to earlier scales that determine the global structural layout.
Existing pruning methods primarily focus on high-frequency detection for token selection, often overlooking structural coherence and consequently degrading global semantics.
To address this limitation, we propose \textbf{StepVAR}, a training-free token pruning framework that accelerates VAR inference by jointly considering structural and textural importance. Specifically, we employ a lightweight high-pass filter to capture local texture details, while leveraging Principal Component Analysis (PCA) to preserve global structural information. This dual-criterion design enables the model to retain tokens critical for both fine-grained fidelity and overall composition. To maintain valid next-scale prediction under sparse tokens, we further introduce a nearest neighbor feature propagation strategy to reconstruct dense feature maps from pruned representations.
Extensive experiments on state-of-the-art text-to-image and text-to-video VAR models demonstrate that StepVAR achieves substantial inference speedups while maintaining generation quality. Quantitative and qualitative evaluations consistently show that our method outperforms existing acceleration approaches, validating its effectiveness and general applicability across diverse VAR architectures.
\end{abstract}

\section{Introduction}
Diffusion models~\cite{ldm, sd3.5, fewshot, expogenius} and AutoRegressive (AR) models~\cite{llamagen, vqgan} have emerged as two dominant paradigms for visual generation. Diffusion models are capable of producing high-fidelity images, yet their inference remains computationally expensive due to the inherently iterative denoising process. In contrast, AR models formulate generation as sequential prediction, offering efficiency--quality trade-off. However, conventional raster-scan \textit{next-token prediction} suffers from unidirectional bias and limited parallelism, resulting in suboptimal scalability for high-resolution synthesis.

Recently, Visual AutoRegressive (VAR) models~\cite{var} introduce a hierarchical alternative by adopting a \textit{next-scale prediction} strategy. Instead of generating tokens in a strict raster order, VAR models progressively predict entire feature maps from coarse to fine resolutions, enabling substantially improved parallelism. This coarse-to-fine design allows VAR to significantly accelerate inference compared to diffusion models and traditional AR approaches, while maintaining competitive generation quality.

Despite its hierarchical design, the computational complexity of VAR still grows quadratically with the number of tokens at higher resolutions. This issue becomes especially critical in high-resolution text-to-image and text-to-video generation tasks~\cite{infinitystar}, where the late stages dominate the overall inference cost. Since VAR is built upon Transformer architectures, the self-attention mechanism incurs a complexity of $\mathcal{O}(N^2)$ with respect to the token number $N$. As the spatial resolution expands progressively across scales, the token count grows substantially, triggering a rapid surge in computational overhead. Consequently, the later fine-scale stages, although primarily responsible for refining local details, contribute disproportionately to the total inference latency.

\begin{figure*}[t]
    \centering
    \begin{minipage}[b]{0.45\linewidth}
        \begin{subfigure}[b]{\linewidth}
            \centering
            \includegraphics[width=\linewidth]{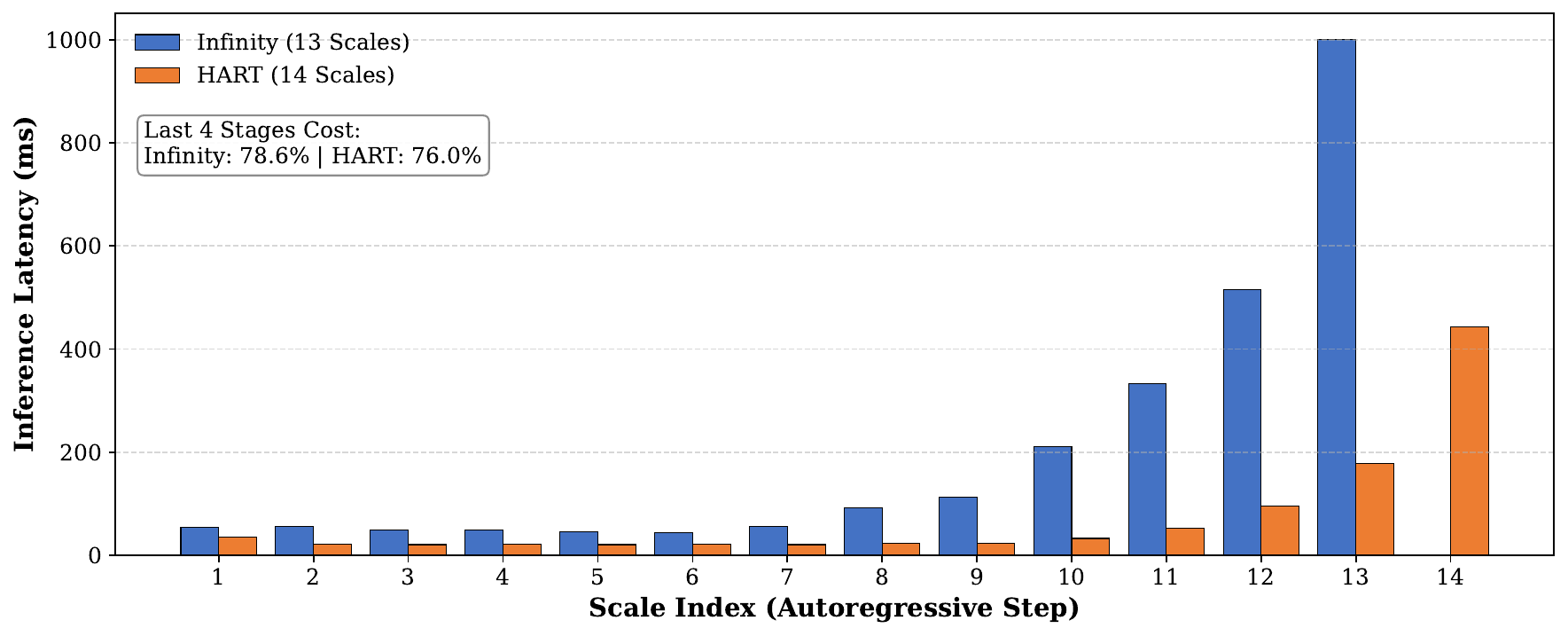}
            \caption{Per-scale latency analysis.}
            \label{fig:latency}
        \end{subfigure}
        \vspace{-.5em}
        \begin{subfigure}[b]{\linewidth}
            \centering
            \includegraphics[width=\linewidth]{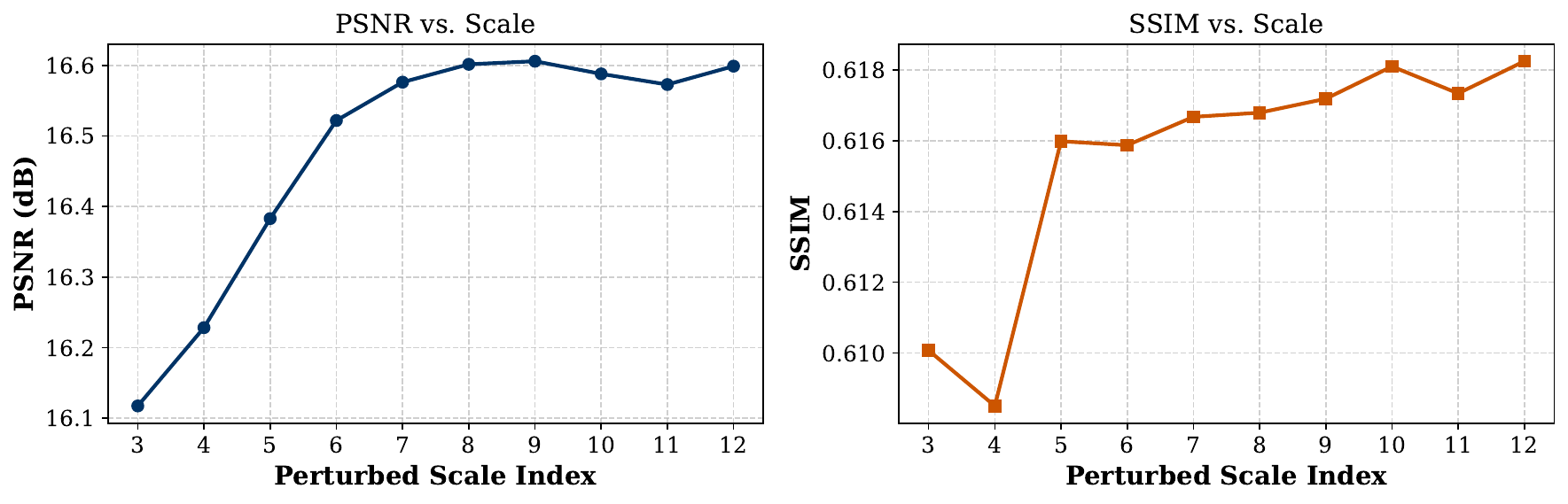}
            \caption{Per-scale Sensitivity Analysis.}
            \label{fig:sensitivity}
        \end{subfigure}
    \end{minipage}
    \hfill
    \begin{minipage}[b]{0.5\linewidth}
        \begin{subfigure}[b]{\linewidth}
            \centering
            \includegraphics[width=\linewidth]{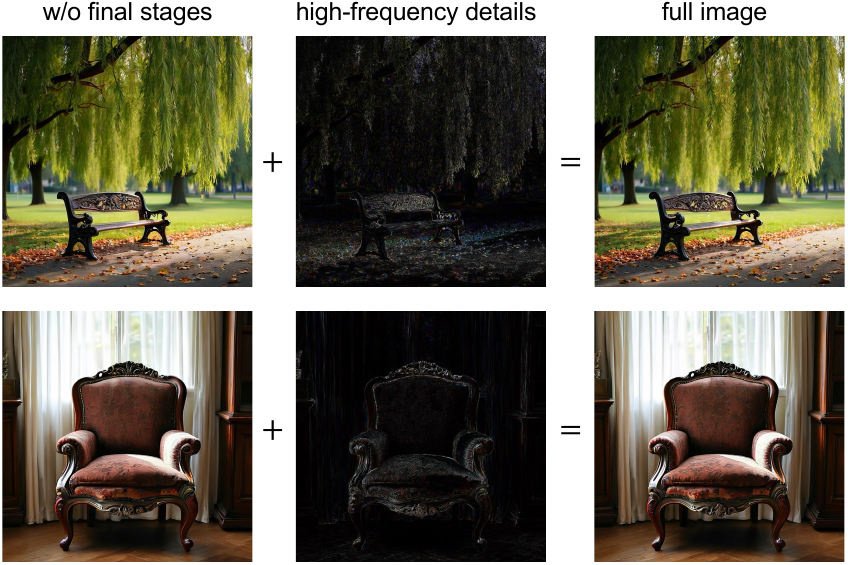} 
            \caption{Visual comparison of model outputs with and without the final autoregressive stages in VAR~\cite{infinity}.}
            \label{fig:freq_vis}
        \end{subfigure}
    \end{minipage}
    
    \caption{\textbf{Inference Characteristics of Visual Autoregressive Models.} (a) The computational bottleneck is heavily skewed towards the final resolution scales, which account for the majority of latency. (b) Sensitivity analysis reveals that early scales are highly sensitive to structural perturbations, whereas later scales exhibit high robustness and error tolerance. (c) Visualizing the residual difference between stages demonstrates that the final scales primarily serve to refine high-frequency textural details.}
    \label{fig:combined_analysis}
\end{figure*}

To address this efficiency challenge, we conduct a systematic analysis of the VAR inference process, as illustrated in Figure~\ref{fig:combined_analysis}, including three key empirical observations. 
First, the computational cost is highly imbalanced across scales. As shown in Figure~\ref{fig:latency}, the final high-resolution stages account for more than 75\% of the total inference time, indicating that latency is dominated by late-scale processing. 
Second, the \textit{next-scale prediction} paradigm exhibits a clear hierarchy of importance. Our sensitivity analysis in Figure~\ref{fig:sensitivity} reveals that perturbations at early coarse scales result in substantial degradation in PSNR and SSIM, whereas later fine scales demonstrate significantly higher tolerance to noise. This suggests that early stages are critical for establishing global structure, while later stages primarily refine details. 
Third, this tolerance is closely related to the spectral characteristics of the generated features. As visualized in Figure~\ref{fig:freq_vis}, the residual differences between outputs with and without the final stages concentrate on high-frequency components, indicating that these computationally intensive steps mainly enhance local textures. 
Collectively, these observations suggest that \textit{while early stages must be strictly preserved to maintain structural integrity, the computationally heavy later stages are highly redundant and serve as prime candidates for token pruning}.

Token pruning is a practical strategy to accelerate inference. Existing pruning methods for generative models, such as FastVAR~\cite{fastvar}, primarily focus on the frequency perspective, identifying important tokens solely based on high-frequency components. While this aligns with the observation that later stages refine details, we argue that \textit{relying exclusively on frequency is insufficient}. Structural information~(principal component), which provides global context for attention calculation, is equally critical. A token may be visually smooth, causing it to be discarded by these frequency-based methods, yet it remains statistically significant for maintaining the global attention routing. Neglecting this structural dimension often leads to semantic drift or artifacts in the generated output.

In this work, we propose \textbf{\methodNAME}, a training-free token pruning method designed to accelerate VAR. Our approach addresses the limitations of purely frequency-based pruning by introducing a \textit{dual-criteria importance criterion}. Specifically, we combine a high-pass filter to capture the necessary high-frequency details with Principal Component Analysis (PCA) to identify tokens that contribute most to the global feature variance, preserving both the textural details and the principal components required for structural coherence. 
To enable seamless next-scale prediction, we further employ a nearest neighbor recovering strategy. By leveraging the high spatial redundancy inherent in feature maps, we propagate information from unpruned tokens to their pruned neighbors, ensuring a dense and semantically consistent input for the subsequent generation scale.

We evaluate our method on representative text-to-image as well as text-to-video models. Quantitative results show that \methodNAME consistently outperforms existing methods across various evaluation metrics. Our contributions can be summarized as follows:
\begin{itemize}
\item We propose a training-free pruning framework \methodNAME, by identifying the uneven computational distribution and hierarchical redundancy in the \textit{next-scale prediction} process.
\item We introduce a dual-criteria pruning mechanism that jointly considers structural and textural importance, accompanied by a nearest neighbor feature propagation strategy that leverages spatial redundancy to ensure consistent recovery.
\item Extensive experiments on both image and video generation tasks demonstrate that our method achieves superior speed-quality trade-offs compared to state-of-the-art acceleration techniques.
\end{itemize}

\section{Related Work}

\subsection{Visual Autoregressive Modeling}
Autoregressive (AR) models have become a cornerstone of visual generation by treating images as sequences of discrete tokens and performing \textit{next-token prediction}. Early approaches like VQVAE~\cite{vqvae} and VQGAN~\cite{vqgan} utilize vector quantization to map images to codebook indices, enabling transformer-based generation. Subsequent works, including LlamaGen~\cite{llamagen} and MAGVIT2~\cite{openmagvit2}, scale this paradigm using advanced language modeling techniques to enhance synthesis quality. Concurrently, models such as MAR~\cite{mar} and NOVA~\cite{nova} improve fidelity by modeling continuous latents directly via diffusion loss, effectively removing the quantization step.

To address the limitations of traditional raster-scan prediction, \textit{Visual Autoregressive (VAR)}\cite{var} models reconceptualize generation via a \textit{next-scale} strategy, synthesizing feature maps in a parallel, coarse-to-fine manner that preserves spatial locality. This hierarchical approach has been rapidly adopted for high-fidelity text-to-image generation\cite{infinity, hart, switti} and extended to the temporal dimension for video synthesis~\cite{infinitystar}, achieving performance competitive with state-of-the-art diffusion models. Beyond generation, the architecture's versatility is demonstrated in pixel-to-pixel tasks like super-resolution~\cite{varsr}, image restoration~\cite{varformer}, and unified understanding-generation frameworks~\cite{onecat, vargpt}, establishing VAR as a robust foundation for diverse visual applications.

\subsection{Acceleration for Visual Generation}
To mitigate the substantial inference latency of diffusion models, diverse acceleration strategies have been proposed. One primary direction focuses on reducing the number of sampling steps through advanced numerical solvers~\cite{dpmsolver, unipc} or knowledge distillation techniques~\cite{dmd, add}. Another avenue optimizes architectural efficiency by leveraging feature caching or skipping redundant layer computations~\cite{deepcache, fasterdiffusion}. Furthermore, token reduction methods exploit spatial and temporal redundancy by dynamically merging or pruning tokens during inference~\cite{tome, toca, dydit, jenga}, effectively lowering the computational cost per step without altering the model weights.

In the realm of VAR models, acceleration strategies are evolving to address the quadratic bottleneck of high-resolution scales. CoDe~\cite{code} employs a collaborative decoding strategy where the large model generates low-frequency content at lower scales, while a smaller specialized model refines high-frequency details at larger scales. FreqExit~\cite{freqexit} introduces a frequency-guided early-exit mechanism for class-to-image VAR models. ScaleKV~\cite{scalekv} focuses on memory constraints through scale-aware KV cache compression, achieving significant memory reduction alongside inference speedups. FastVAR~\cite{fastvar} and SparseVAR~\cite{sparsevar} propose training-free pruning strategies that prioritize high-frequency tokens, leveraging the observation that later generation stages are dominated by textural refinement. While these methods effectively exploit frequency priors, our framework further enhances this paradigm by explicitly accounting for both high-frequency details and global structural pivots, ensuring that essential context is preserved during token reduction.

\section{Method}

\begin{figure*}[t]
\begin{center}
\includegraphics[width=\linewidth]{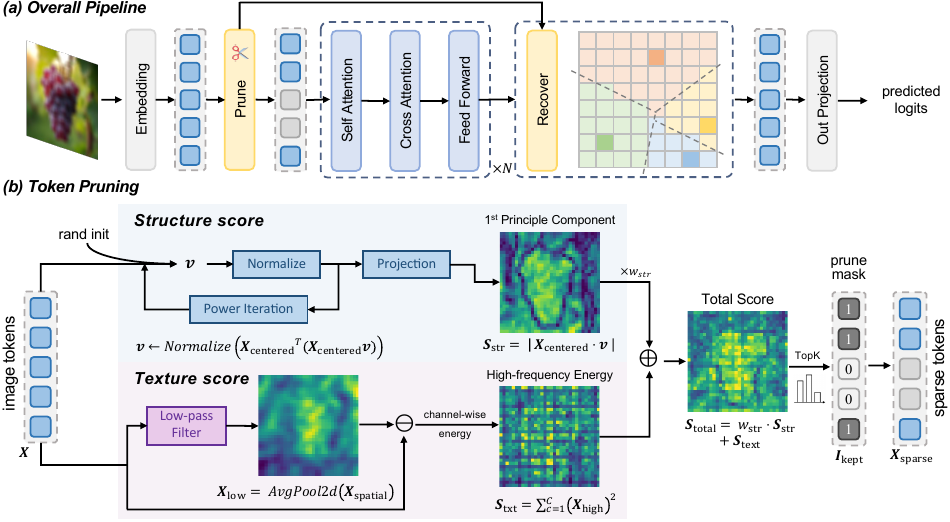}
\end{center}
\caption{\textbf{Overview of the \methodNAME Framework.} (a) \textbf{Overall Pipeline:} Input tokens are pruned prior to entering the transformer blocks. In the \textbf{Recover} stage, the dense feature map required for the next scale is reconstructed using Nearest Neighbor Feature Propagation. (b) \textbf{Token Pruning:} Our Structure-Texture Guided strategy calculates a joint importance score by weighting a Structural Score ($\mathbf{S}_{\text{str}}$), derived from PCA analysis, and a Textural Score ($\mathbf{S}_{\text{txt}}$), derived from spatial high-frequency filter. This ensures the retention of tokens critical for both global layout and local details.}
\label{fig:method}
\end{figure*}

In this section, we detail the design of \textbf{\methodNAME}, a framework tailored for accelerating Visual Autoregressive (VAR) models. We begin by analyzing the specific computational bottlenecks in the next-scale prediction paradigm. We then introduce our novel token selection strategy, which operates on a dual-criteria basis: leveraging Principal Component Analysis (PCA) to preserve structural integrity and high-pass filtering to retain textural fidelity. Finally, we present our feature propagation mechanism, which reconstructs the full-resolution feature map from sparse tokens using a nearest neighbor approach, ensuring semantic continuity. 

\subsection{Preliminaries}

The core idea of the VAR framework lies in its departure from the raster-scan order of standard autoregressive models. Instead of predicting a single token $x_t$ based on the history $x_{<t}$, VAR generates the image latent $Z$ through modeling a sequence of scales $(z_1, z_2, \dots, z_K)$, in which $z_s \in \mathbb{R}^{h_s \times w_s \times C}$ represents the entire feature map at resolution $h_s \times w_s$. Then the overall objective is factorized autoregressively:
\begin{equation}
    p(Z) = \prod_{s=1}^{K} p(z_s | z_{<s}).
\end{equation}
During the generation of scale $s$, the model attends to all tokens within the current feature map $z_s$ and the cached features of previous scales.

While enabling intra-scale parallelism, the next-scale paradigm introduces severe computational asymmetry. With resolution typically following a progressive schedule (e.g., $1, \dots, 64$), the sequence length $L_s$ expands significantly in the final stages, triggering the large $\mathcal{O}(L_s^2)$ complexity of self-attention. Consequently, as shown in Figure~\ref{fig:latency}, the inference cost is heavily skewed, with the final few high-resolution stages dwarfing the combined cost of all preceding ones. Leveraging our observation that these expensive stages are highly redundant (Figure~\ref{fig:combined_analysis}), our objective is to reduce the effective sequence length $L_s$ to a subset $k \ll L_s$ specifically in these later steps. Critically, this reduction applies systemically across the entire transformer including self-attention, cross-attention, and FFN, thereby maximizing latency and memory savings without the need for retraining.

\subsection{Structure-Texture Guided Token Pruning}

The efficacy of token pruning relies heavily on the selection strategy: determining which tokens are redundant requires a metric that aligns with the model's generative objectives. Prior training-free approaches have predominantly relied on high-frequency metrics, operating on the assumption that only high-frequency or texture-rich regions require active computation in later stages. However, we argue that this view is incomplete. A comprehensive pruning strategy for VAR models should preserve two distinct categories of information:
\begin{enumerate}
    \item \textbf{Structural Integrity:} Features that maintain global layout and semantic consistency. These are often encoded in \textit{low-frequency} tokens that serve as critical anchors for the attention mechanism.
    \item \textbf{Textural Refinement:} Features that are responsible for \textit{high-frequency} details, such as edges, intricate patterns, and sharp boundaries.
\end{enumerate}
To address this, \methodNAME employs a dual-criteria scoring mechanism that explicitly evaluates both the structural and textural importance of each token.

\noindent \textbf{Structural Integrity Score.}
To ensure semantic coherence, it is essential to preserve tokens that maintain the global layout and composition of the image. Tokens representing broad or homogeneous regions, such as backgrounds and smooth gradients, often encode vital structural information and serve as global attention anchors. Neglecting them severs the global context, causing the model to lose track of the overall composition.

\begin{figure}[t]
\centering
\includegraphics[width=0.6\linewidth]{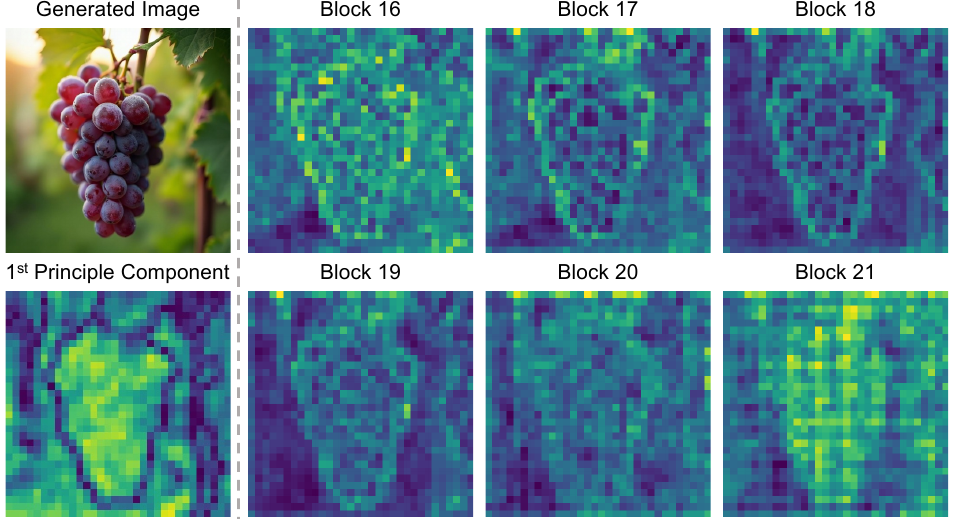}
\caption{\textbf{Correlation of first principle component and transformer block residual.}}
\label{fig:pca_visualize}
\end{figure}

We hypothesize that the tokens most critical for structural processing are those with the highest magnitude along the principal feature directions. To validate this, we conducted an empirical analysis comparing the \textit{block residual} (the Mean Squared Error between a transformer block's input and output) with the first principle component, as shown in Fig.~\ref{fig:pca_visualize}. We observed a strong spatial correlation: the regions where the transformer network modifies the features most aggressively correspond to the tokens with high projections on the principal components. This suggests that keeping tokens with high PCA scores allows us to retain the exact subset of features the network intends to manipulate.

Since performing Singular Value Decomposition (SVD) is too costly for real-time inference ($\mathcal{O}(LC^2)$), we employ the \textit{Power Iteration} method to approximate the first principal component. Formally, let $\mathbf{X} \in \mathbb{R}^{B \times L \times C}$ be the input feature map. We first calculate the centered features $\mathbf{X}_{\text{centered}} = \mathbf{X} - \bar{\mathbf{X}}$, then initialize a random vector $v \in \mathbb{R}^{C \times 1}$ and iteratively update it:
\begin{equation}
    \mathbf{v} \leftarrow \frac{\mathbf{X}_{\text{centered}}^T (\mathbf{X}_{\text{centered}} \mathbf{v})}{||\mathbf{X}_{\text{centered}}^T (\mathbf{X}_{\text{centered}} \mathbf{v})||_2}.
\end{equation}
In practice, we find that $T=3$ to $5$ iterations are sufficient for convergence. The structural importance score $S_{\text{str}}$ is the magnitude of the projection of each token onto $\mathbf{v}$:
\begin{equation}
    \mathbf{S}_{\text{str}} = | \mathbf{X}_{\text{centered}} \cdot \mathbf{v} |.
\end{equation}
This score identifies statistically significant tokens regardless of their spatial frequency, effectively protecting the global structure from being pruned.

\noindent \textbf{Textural Refinement Score.}
In the hierarchical generation process of VAR models, the role of features evolves across scales. While early scales establish the global composition, the later stages, where the computational bottleneck is most severe, are primarily responsible for high-frequency refinement. These stages populate the image with intricate details such as texture gradients and sharp object boundaries. Consequently, tokens corresponding to these high-frequency regions carry the majority of the information introduced at that specific scale.

\begin{algorithm}[t]
\caption{Structure-Texture Guided Token Pruning.}
\label{alg:selection}
\begin{algorithmic}[1]
\REQUIRE Input features $\mathbf{X} \in \mathbb{R}^{B \times L \times C}$, Spatial dimensions $H, W$, Pruning ratio $r \in [0, 1]$, PCA weight $w_{\text{str}}$ (default: 0.5), Power iterations $T$ (default: 3).
\STATE \textbf{// 1. Compute Structural Score (PCA)}
\STATE $\mathbf{X}_{\text{centered}} \leftarrow \mathbf{X} - \text{Mean}(\mathbf{X}, \text{dim}=1);$
\STATE Initialize random unit vector $\mathbf{v} \in \mathbb{R}^{B \times C \times 1}$;
\FOR{$t \leftarrow 1$ \TO $T$}
    \STATE $\mathbf{v} \leftarrow \text{Normalize}(\mathbf{X}_{\text{centered}}^T (\mathbf{X}_{\text{centered}} \mathbf{v}));$
\ENDFOR
\STATE $\mathbf{S}_{\text{str}} \leftarrow |\mathbf{X}_{\text{centered}} \cdot \mathbf{v}|$;

\STATE \textbf{// 2. Compute Textural Score (High-Pass)}
\STATE $\mathbf{X}_{\text{spatial}} \leftarrow \text{Reshape}(\mathbf{X}, (B, C, H, W))$;
\STATE $\mathbf{X}_{\text{low}} \leftarrow \text{AvgPool2d}(\mathbf{X}_{\text{spatial}})$;
\STATE $\mathbf{X}_{\text{high}} \leftarrow \mathbf{X}_{\text{spatial}} - \mathbf{X}_{\text{low}}$;
\STATE $\mathbf{S}_{\text{txt}} \leftarrow \sum_{c=1}^{C} (\mathbf{X}_{\text{high}})^2$;

\STATE \textbf{// 3. Joint Selection}
\STATE $\mathbf{S}_{\text{total}} \leftarrow w_{\text{str}} \cdot \mathbf{S}_{\text{str}} + \mathbf{S}_{\text{txt}}$;
\STATE $k \leftarrow \lfloor (1-r) \cdot L \rfloor$;
\STATE $\mathcal{I}_{\text{kept}} \leftarrow \text{TopK}(\mathbf{S}_{\text{total}}, k)$;
\STATE $\mathbf{X}_{\text{sparse}} \leftarrow \text{Gather}(\mathbf{X}, \mathcal{I}_{\text{kept}})$;

\RETURN $\mathbf{X}_{\text{sparse}}, \mathcal{I}_{\text{kept}}$.
\end{algorithmic}
\end{algorithm}

To identify these regions of rapid spatial change without the overhead of spectral transforms like the Fast Fourier Transform, we adopt a computationally efficient spatial approximation. We premise that high-frequency tokens, such as those representing edges or texture, exhibit significant deviation from their local neighborhood averages. In contrast, tokens in flat regions remain close to the local mean.
Formally, we first reshape the input feature $\mathbf{X} \in \mathbb{R}^{B \times L \times C}$ into $\mathbf{X}_{\text{spatial}} \in \mathbb{R}^{B \times C \times H \times W}$. We then compute the local low-frequency approximation $\mathbf{X}_{\text{low}}$ using a $3 \times 3$ average pooling operation with padding to preserve dimensions:
\begin{equation}
    \mathbf{X}_{\text{low}} = AvgPool2d(\mathbf{X}_{\text{spatial}}).
\end{equation}
The high-frequency component, $\mathbf{X}_{\text{high}}$, is obtained as the residual difference between the original signal and its smoothed version:
\begin{equation}
    \mathbf{X}_{\text{high}} = \mathbf{X}_{\text{spatial}} - \mathbf{X}_{\text{low}}.
\end{equation}
We then condense this multi-channel residual into a scalar importance score $S_{\text{txt}}$ by computing the energy ($L_2$ norm) across the channel dimension:
\begin{equation}
    \mathbf{S}_{\text{txt}} = \sum_{c=1}^{C} (\mathbf{X}_{\text{high}}^{(c)})^2.
\end{equation}
This metric provides a simple yet robust measure of textural importance. Crucially, it relies solely on elementary tensor operations such as pooling and subtraction, rendering its computational cost negligible compared to the quadratic attention mechanism it accelerates. By prioritizing tokens with high $\mathbf{S}_{\text{txt}}$, we ensure that the model retains the capacity to generate sharp edges and complex textures, preventing the over-smoothing artifacts common in aggressive pruning strategies.

\noindent \textbf{Joint Selection}
The final importance score $S_\text{total}$ is a weighted fusion of the two metrics:
\begin{equation}
    \mathbf{S}_{\text{total}} = w_{\text{str}} \cdot \mathbf{S}_{\text{str}} + \mathbf{S}_{\text{txt}}.
\end{equation}
We select the top-$k$ tokens with the highest $\mathbf{S}_\text{total}$ values, where $k$ is determined by the pruning ratio $r$. The indices of these tokens, $\mathcal{I}_\text{kept}$, form the sparse sequence passed to the subsequent transformer layers. We summarize our token pruning algorithm in Algorithm~\ref{alg:selection}.

\subsection{Nearest Neighbor Feature Propagation}

\begin{algorithm}[t]
\caption{Nearest-Neighbor Feature Propagation.}
\label{alg:recovery}
\begin{algorithmic}[1]
\REQUIRE Sparse features $\mathbf{X}'_{\text{sparse}} \in \mathbb{R}^{B \times k \times C}$, Kept indices $\mathcal{I}_{\text{kept}} \in \mathbb{R}^{B \times k}$, Spatial dimensions $H, W$.
\STATE \textbf{// 1. Define Spatial Coordinates}
\STATE $\mathcal{G}_{\text{all}} \leftarrow \text{MeshGrid}((0, \dots, H-1), (0, \dots, W-1))$;
\STATE $\mathcal{G}_{\text{kept}} \leftarrow \text{Gather}(\mathcal{G}_{\text{all}}, \mathcal{I}_{\text{kept}})$;

\STATE \textbf{// 2. Find Nearest Neighbors}
\STATE $\mathbf{D} \leftarrow \text{PairwiseDistance}(\mathcal{G}_{\text{all}}, \mathcal{G}_{\text{kept}})$;
\STATE $\text{idx}_{\text{nn}} \leftarrow \arg\min(\mathbf{D}, \text{dim}=-1)$;

\STATE \textbf{// 3. Propagate Features}
\STATE $\mathbf{X}_{\text{rec}} \leftarrow \text{Gather}(\mathbf{X}'_{\text{sparse}}, \text{idx}_{\text{nn}})$;

\RETURN $\mathbf{X}_{\text{rec}}$.
\end{algorithmic}
\end{algorithm}

Once the sparse tokens $\mathbf{X}_{\text{sparse}}$ have been processed by the transformer blocks of the current scale, we obtain an updated sparse representation $\mathbf{X}'_{\text{sparse}}$. However, the VAR paradigm requires a full dense feature map to serve as the condition for the next scale prediction. The challenge lies in reconstructing the full $H \times W$ grid from the $k$ updated tokens.

Standard approaches typically copy from designated anchor tokens or re-use cached features from the previous scale. We argue that both approaches are suboptimal: copying from anchor tokens ignores the spacial structure, and could disrupt the spatial structure of the feature map, while utilizing previous scale features creates a semantic gap, as the features have not benefited from the current scale's processing.

To address this, we propose \textbf{Nearest Neighbor Feature Propagation}. This method exploits the high spatial redundancy of the latent feature. We posit that the feature value of a pruned token is best approximated by the feature value of its spatially closest kept token within the \textit{current} scale.

Mathematically, let $\mathcal{G}_{\text{all}}$ be the set of $(y, x)$ coordinates for the full grid, and $\mathcal{G}_{\text{kept}}$ be the coordinates corresponding to $\mathcal{I}_{\text{kept}}$. For every position $p_i \in \mathcal{G}_{\text{all}}$ that was pruned, we identify the nearest active neighbor $p_j \in \mathcal{G}_{\text{kept}}$:
\begin{equation}
    j^* = \arg\min_{j} || p_i - p_j ||_2.
\end{equation}
The reconstructed feature at position $i$ is then assigned the value of the processed token at index $j^*$: $\mathbf{X}_{\text{rec}}[i] = X'_{\text{sparse}}[j^*]$.

Functionally, this operation performs an adaptive nearest neighbor interpolation, where each kept token fills the gaps in its immediate spatial vicinity. This ensures that: (1) every position in the reconstructed map is populated with a valid, processed feature from the current semantic level, avoiding the domain shift associated with copying from previous scales; and (2) the spatial structure is respected across both detailed and flat regions. In high-frequency areas, our pruning strategy maintains high token density to preserve detail, while in flat regions, the high spatial redundancy allows pruned tokens to be accurately approximated by their neighbors. Unlike copying from arbitrary anchor tokens, which ignores local spatial relationships, this mechanism (Algorithm~\ref{alg:recovery}) effectively bridges the sparsity of the current stage with the density required for the next, preserving the semantic integrity of the feature map.

\section{Experiments}

\subsection{Experimental Setup}

\noindent \textbf{Models and Datasets.}
We evaluate the effectiveness of \methodNAME across representative high-resolution image and video generation models. For text-to-image (T2I) tasks, we employ Infinity~\cite{infinity} and HART~\cite{hart}, both capable of generating images up to $1024 \times 1024$ resolution. For text-to-video (T2V) generation, we utilize InfinityStar~\cite{infinitystar}, which extends the scale-wise paradigm to the temporal dimension.

To comprehensively assess generation quality, we prioritize semantic consistency metrics. We report GenEval~\cite{geneval} and DPG~\cite{ella} scores to evaluate high-level semantic consistency and detail preservation for complex prompts. Additionally, we utilize the MJHQ-30K~\cite{playgroundv2.5} benchmark to compute standard FID~\cite{fid} and CLIP Scores \cite{CLIP-Score}, assessing perceptual fidelity and text-image alignment. For T2V evaluation, we report results on VBench~\cite{vbench}, a comprehensive benchmark measuring video quality, temporal consistency, and motion dynamics.

\noindent \textbf{Pruning Configuration.}
Our method is training-free and applied directly to the pre-trained checkpoints. To ensure a fair and direct comparison with the baseline method FastVAR~\cite{fastvar}, we adopt consistent pruning configurations for the corresponding models. For Infinity, we apply pruning to the last 4 scale steps with ratios of $\{40\%, 50\%, 100\%, 100\%\}$. The $100\%$ pruning ratio indicates that the generative step is skipped entirely. For HART, we prune the last 2 scales with ratios of $\{50\%, 75\%\}$. For the video model InfinityStar, we apply pruning to the final stages with ratios of $\{50\%, 100\%\}$.

\noindent \textbf{Performance Measurement.}
We measure the inference efficiency by recording the end-to-end latency, which encompasses the text encoder, the autoregressive transformer backbone, and the VAE decoder. We report the average inference time using a fixed set of prompts to ensure consistency across all methods. Text-to-image experiments are conducted on a single NVIDIA RTX 3090 GPU (24GB), while text-to-video experiments are performed on an NVIDIA A40 GPU (48GB) to accommodate the larger memory requirements.

\begin{figure*}[t]
\begin{center}
\includegraphics[width=\linewidth]{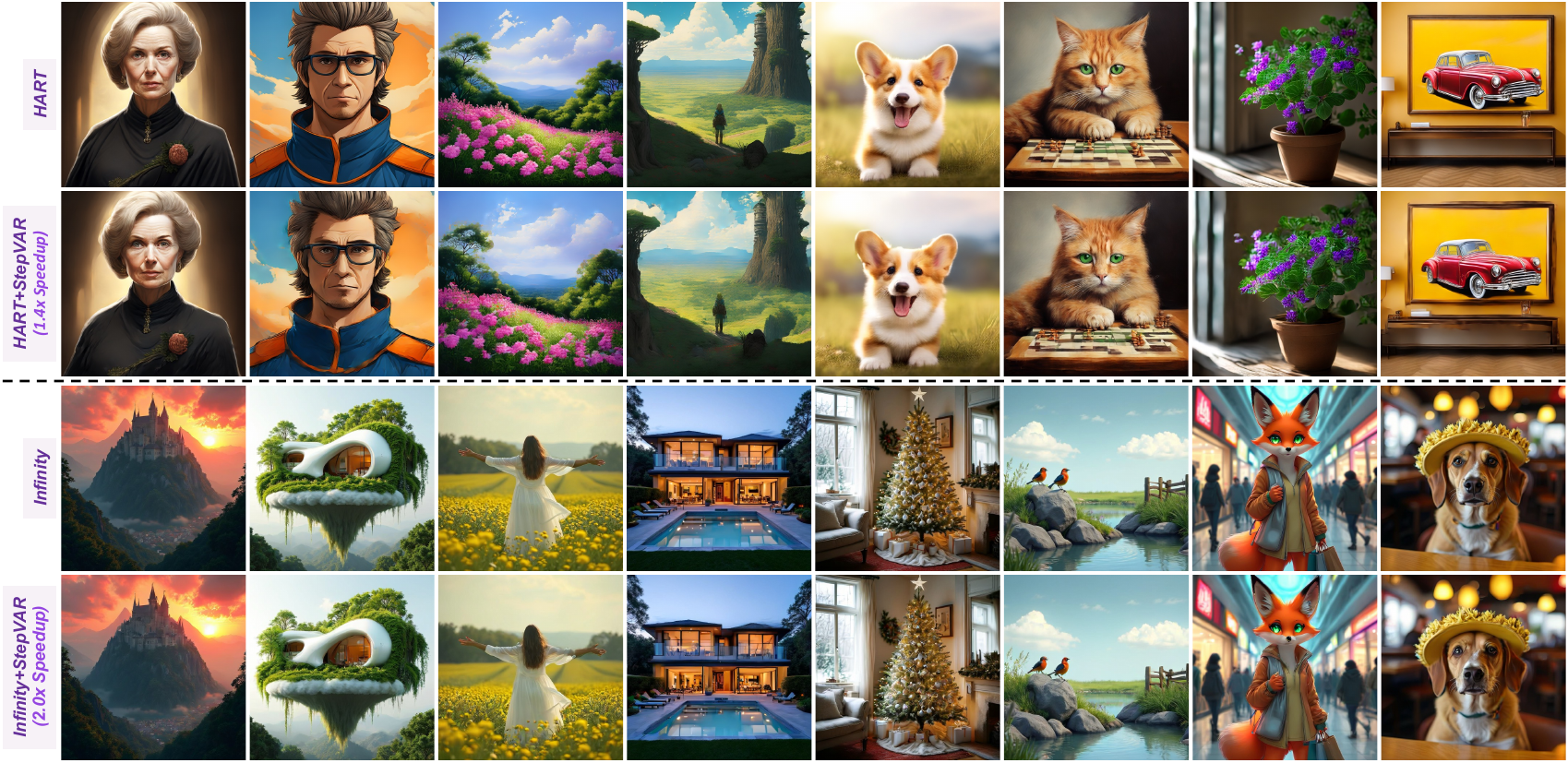}
\end{center}
\caption{\textbf{Visual Comparison between original model and our \methodNAME generated images.}}
\label{fig:visual_comparison}
\end{figure*}

\begin{table*}[t]
\centering
\captionsetup{skip=5pt} 
\caption{Evaluation on the GenEval and DPG benchmark.}\label{tab:genEvalSotaTable}
\resizebox{\linewidth}{!}{ 
\begin{tabular}{lcccccccccc}
    \toprule
    \multirow{2}{*}{Methods} & \multicolumn{3}{c}{Inference Efficiency} & \multicolumn{4}{c}{GenEval$\uparrow$} & \multicolumn{3}{c}{DPG$\uparrow$} \\
    \cmidrule(lr){2-4}\cmidrule(lr){5-8}\cmidrule(l){9-11}
    & \# Params & SpeedUp & Latency & Two Obj. & Position & Color Attri. & Overall & Global & Relation & Overall \\
    \midrule
    PixArt-alpha~\cite{pixartalpha} & 0.6B & - & - & 0.50 & 0.08 & 0.07 & 0.48 & 74.97 & 82.57 & 71.11 \\
    SDXL~\cite{sdxl} & 2.6B & - & - & 0.74 & 0.15 & 0.23 & 0.55 & 83.27 & 86.76 & 74.65 \\
    SD3~\cite{sd3.5} & 8B & - & - & 0.89 & 0.34 & 0.47 & 0.71 & - & - & - \\
    LlamaGen~\cite{llamagen} & 0.8B & - & - & 0.34 & 0.07 & 0.04 & 0.32 & & & 65.16 \\
    Show-o~\cite{showo} & 1.3B & - & - & 0.80 & 0.31 & 0.50 & 0.68 & - & - & 67.48 \\
    \midrule
    \textit{HART}~\cite{hart} & 732M & 1.0$\times$ & 1.40s & 0.49 & 0.16 & 0.18 & 0.51 & 84.77 & 84.87 & 74.87 \\
    +SparseVAR~\cite{sparsevar} & 732M & 1.4$\times$ & 1.01s & 0.59 & 0.10 & 0.20 & 0.50 & 85.41 & 89.12 & 74.45 \\
    +FastVAR~\cite{fastvar} & 732M & 1.4$\times$ & 1.03s & 0.56 & 0.12 & 0.20 & 0.51 & 80.44 & 85.46 & 74.18  \\
    \rowcolor{green!5} \textbf{+Ours} & 732M & 1.4$\times$ & 0.98s & 0.60 & 0.10 & 0.19 & 0.51 & 83.29 & 80.42 & 74.58  \\
    \midrule
    \textit{Infinity}~\cite{infinity} & 2B & 1.0$\times$ & 3.27s & 0.85 & 0.49 & 0.57 & 0.73 & 90.13 & 89.18 & 82.72 \\
    +SparseVAR~\cite{sparsevar} & 2B & 2.0$\times$ & 1.67s & 0.85 & 0.42 & 0.55 & 0.72 & 85.11 & 92.46 & 82.56 \\
    +FastVAR~\cite{fastvar} & 2B & 2.0$\times$ & 1.60s & 0.83 & 0.42 & 0.57 & 0.72 & 88.09 & 84.19 & 82.20 \\
    \rowcolor{green!5} \textbf{+Ours} & 2B & 2.0$\times$ & 1.60s & 0.83 & 0.44 & 0.57 & 0.72 & 84.52 & 90.97 & 82.65 \\
    \bottomrule
\end{tabular}
}
\end{table*}

\subsection{Main Results}

\noindent \textbf{Text-to-Image Generation Quality.}
We first evaluate semantic consistency and detail preservation on the GenEval and DPG benchmarks, comparing \methodNAME against SparseVAR~\cite{sparsevar} and FastVAR~\cite{fastvar}. As shown in Table~\ref{tab:genEvalSotaTable}, our method demonstrates consistent superiority across architectures. On HART ($1.4\times$ speedup), \methodNAME achieves an Overall GenEval of 0.51 and DPG of 74.58, effectively matching the original model and surpassing both baselines. On Infinity ($2.0\times$ speedup), \methodNAME maintains robustness with an Overall DPG of 82.65, surpassing FastVAR (82.20) and SparseVAR (82.56). These results validate that guiding pruning with both structural and textural criteria effectively mitigates quality degradation compared to methods relying on isolated metrics.

To provide a holistic view of image quality beyond semantic consistency, we assess perceptual fidelity and alignment using the MJHQ-30K dataset. Table~\ref{tab:mjhq30k_eval} reports scores for representative categories and the overall dataset. \methodNAME consistently outperforms the baselines in distribution-level quality. Notably on HART, we achieve an overall FID of 10.28, improving upon FastVAR (10.52) and even the original baseline (11.00), likely due to the suppression of redundant noisy tokens. Similarly, on Infinity, \methodNAME achieves the best overall FID (9.85). Meanwhile, CLIP scores remain stable across the board, ensuring that our pruning strategy maintains strong alignment with the input text prompts.

\begin{table}[t]
\centering
\caption{Evaluation on MJHQ-30K Dataset. We report FID and CLIP Score on selected categories and overall performance.}\label{tab:mjhq30k_eval}
\begin{tabular}{lc cccccccc}
    \toprule
    \multirow{2}{*}{Methods} & \multirow{2}{*}{SpeedUp} & \multicolumn{2}{c}{Food} & \multicolumn{2}{c}{Plants} & \multicolumn{2}{c}{Animals} & \multicolumn{2}{c}{Overall} \\
    \cmidrule(lr){3-4} \cmidrule(lr){5-6} \cmidrule(lr){7-8} \cmidrule(l){9-10}
    & & FID$\downarrow$ & CLIP$\uparrow$ & FID$\downarrow$ & CLIP$\uparrow$ & FID$\downarrow$ & CLIP$\uparrow$ & FID$\downarrow$ & CLIP$\uparrow$ \\
    \midrule
    \textit{HART} & 1.0$\times$ & 30.39 & 0.28 & 31.14 & 0.28 & 22.35 & 0.30 & 11.00 & 0.283 \\
    +FastVAR & 1.4$\times$ & 30.92 & 0.28 & 36.35 & 0.27 & 23.24 & 0.30 & 10.52 & 0.280 \\
    \rowcolor{green!5}\textbf{+Ours} & 1.4$\times$ & 30.28 & 0.28 & 30.95 & 0.28 & 22.00 & 0.30 & 10.28 & 0.283 \\
    \midrule
    \textit{Infinity} & 1.0$\times$ & 32.13 & 0.27 & 30.49 & 0.26 & 25.12 & 0.29 & 10.09 & 0.275 \\
    +FastVAR & 2.0$\times$ & 31.96 & 0.27 & 30.16 & 0.27 & 25.24 & 0.30 & 10.08 & 0.278 \\
    \rowcolor{green!5}\textbf{+Ours} & 2.0$\times$ & 32.09 & 0.27 & 30.35 & 0.27 & 24.78 & 0.30 & 9.85 & 0.278 \\
    \bottomrule
\end{tabular}
\end{table}

\noindent \textbf{Qualitative Comparison.}
Visual comparisons presented in Figure~\ref{fig:visual_comparison} further corroborate our quantitative findings. Even under $1.4\times$ acceleration on HART and $2.0\times$ on Infinity, images generated by \methodNAME are visually indistinguishable from the original outputs. Our structure and texture guided pruning preserves the global layout and structural integrity of complex scenes, ensuring high-fidelity generation despite the reduction in token count.

\noindent \textbf{Text-to-Video Generation Quality.}
Finally, we demonstrate the generalizability of our approach by extending the evaluation to the temporal dimension using InfinityStar. As summarized in Table~\ref{tab:vBenchTable}, \methodNAME achieves a $1.4\times$ speedup with negligible impact on video quality. The Overall VBench score reaches 83.35, closely tracking the original model's 83.88. This confirms that \methodNAME's structure-texture hypothesis extends effectively to video latents, enabling acceleration for computationally intensive video generation tasks.

\subsection{Ablation Studies}

To validate the effectiveness of our proposed mechanisms, we conduct extensive ablation studies on the HART model. We analyze the impact of different token pruning criteria, feature recovery strategies, pruning ratios, and stage selection. Unless otherwise specified, experiments are conducted with a pruning configuration that yields an approximate $1.4\times$ speedup to ensure a fair comparison basis.

\begin{table}[t]
    \centering
    \caption{Evaluation on the VBench benchmark.}
    \label{tab:vBenchTable}
    \begin{tabular}{lcccccc}
   \toprule
        \multirow{2}{*}{Models} & \multirow{2}{*}{\# Params} & \multirow{2}{*}{SpeedUp} & Quality & Semantic & \multirow{2}{*}{Overall}  \\ 
         &  & & Score & Score &  \\
        \midrule
        CogVideoX-5B~\cite{cogvideox} & 5B & - & 82.75 & 77.04 & 81.61 \\
        HunyuanVideo~\cite{hunyuanvideo} & 13B & - & 85.09 & 75.82 & 83.24 \\
        Emu3~\cite{emu3} & 8B & - & 84.09 & 68.43 & 80.96 \\
        \midrule
        InfinityStar~\cite{infinitystar} & 8B & 1.0 $\times$ & 84.01 & 83.38 & 83.88 \\
        \rowcolor{green!5} \textbf{+Ours} & 8B & 1.4 $\times$ & 83.46 & 82.91 & 83.35 \\
        \bottomrule
    \end{tabular}
\end{table}

\noindent \textbf{Impact of Token Pruning Strategy.}
We verify the necessity of our dual-criteria scoring by comparing it against three baselines: (1) \textit{Random pruning}, (2) \textit{L2-norm based pruning}, which uses L2 distance with the mean to estimate importance, similar to FastVAR~\cite{fastvar}, and (3) \textit{High Frequency (HF) only}, which uses our proposed high-pass score without the PCA component. Results are summarized in the upper section of Table~\ref{tab:ablation_combined}.
Random pruning results in a significant drop in both semantic consistency (GenEval 0.48) and detail (DPG 73.86), confirming that token redundancy is not uniform. While the L2-norm and HF-only strategies improve performance by identifying texture-rich regions, the full \textit{HF+PCA} strategy achieves the highest scores across all metrics (GenEval 0.51, DPG 74.58). This validates that incorporating Structural (PCA) information ensures that structural integrity is preserved alongside local textural details.

\begin{table}[htbp]
    \centering
    \begin{minipage}[t]{0.55\textwidth}
        \centering
        \caption{Ablation study on token pruning and recovering strategies. We evaluate the impact of different approaches using GenEval, DPG, FID, and CLIP metrics.}\label{tab:ablation_combined}
        \resizebox{\linewidth}{!}{
        \begin{tabular}{lcccc}
            \toprule
            Strategy & GenEval $\uparrow$ & DPG $\uparrow$ & FID $\downarrow$ & CLIP $\uparrow$ \\
            \midrule
            \multicolumn{5}{l}{\textit{Ablation on Token Pruning Strategy}} \\
            Random & 0.48 & 73.86 & 32.17 & 0.28 \\
            L2-norm~\cite{fastvar} & 0.50 & 74.56 & 31.05 & 0.29 \\
            High Frequency & 0.51 & 74.45 & 29.95 & 0.29 \\ 
            \rowcolor{green!5} \textbf{HF+PCA (Ours)} & \textbf{0.51} & \textbf{74.58} & \textbf{29.84} & \textbf{0.29} \\
            \midrule
            \multicolumn{5}{l}{\textit{Ablation on Token Recovering Strategy}} \\
            Cache~\cite{fastvar} & 0.50 & 74.45 & \textbf{27.37} & 0.28 \\
            Anchor Tokens & 0.50 & 74.14 & 30.60 & 0.29 \\
            \rowcolor{green!5} \textbf{Nearest Neighbor} & \textbf{0.51} & \textbf{74.58} & 29.84 & \textbf{0.29} \\
            \bottomrule
        \end{tabular}
        }
    \end{minipage}
    \hfill
    \begin{minipage}[t]{0.43\textwidth}
        \centering
        \caption{Ablation study on prune ratio. We evaluate the trade-off between efficiency and generation quality.}\label{tab:ablation_prune_ratio}
        \resizebox{\linewidth}{!}{
        \begin{tabular}{lccccc}
            \toprule
            Ratio & SpeedUp & GenEval$\uparrow$ & DPG$\uparrow$ & FID $\downarrow$ & CLIP $\uparrow$ \\
            \midrule
            0 & 1.0$\times$ & 0.51 & 73.77 & 30.45 & 0.28 \\ 
            0.3 & 1.20$\times$ & 0.50 & 74.89 & 30.24 & 0.29 \\ 
            0.5 & 1.32$\times$ & 0.51 & 74.81 & 29.91 & 0.29 \\ 
            0.7 & 1.43$\times$ & 0.50 & 74.10 & 32.53 & 0.28 \\ 
            0.9 & 1.52$\times$ & 0.49 & 73.36 & 34.11 & 0.28 \\ 
            \bottomrule
        \end{tabular}
        }
    \end{minipage}
\end{table}

\begin{figure}[t]
\begin{center}
\includegraphics[width=0.7\linewidth]{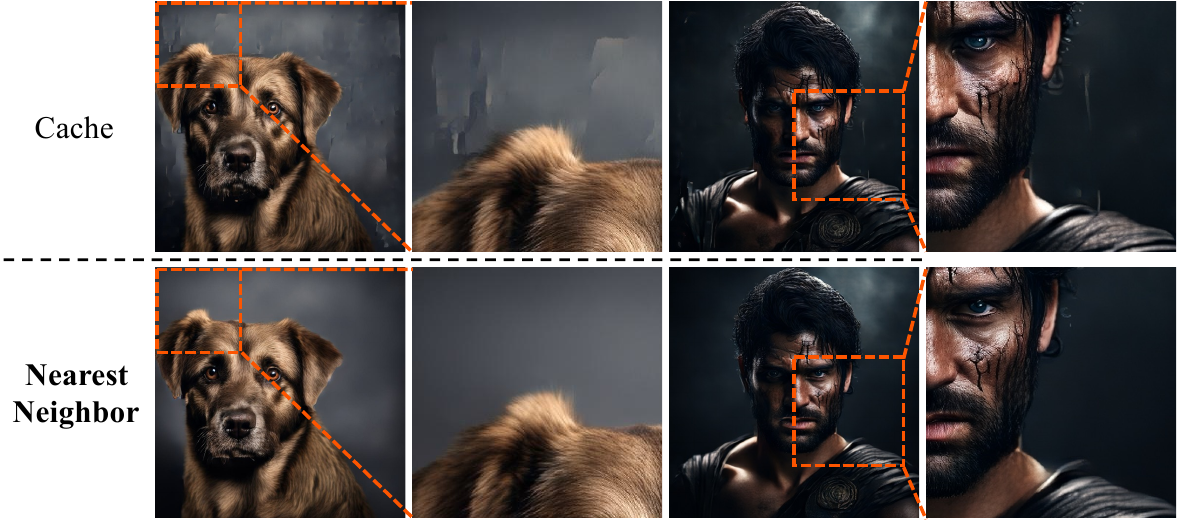}
\end{center}
\caption{\textbf{Visual comparison of different recovery strategies.}}
\label{fig:recover_visualization}
\end{figure}

\noindent \textbf{Impact of Recovery Strategy.}
Next, we investigate how to best reconstruct the dense feature map required for the next-scale prediction. We compare three approaches: (1) \textit{Cache}, which fills pruned locations by upsampling the feature map from the previous scale; (2) \textit{Anchor Tokens}, which forces the retention of a fixed grid of tokens (stride 3) to serve as copy sources; and (3) our proposed \textit{Nearest Neighbor Feature Propagation}, which propagates features from the spatially nearest kept token in the current scale.
As shown in Table~\ref{tab:ablation_combined}, the \textit{Nearest Neighbor Feature Propagation} strategy delivers the best semantic consistency and text alignment. While the \textit{Cache} strategy yields a numerically lower FID (27.37), qualitative results as visualized in Figure~\ref{fig:recover_visualization} show that images generated using the Cache strategy exhibit noticeable artifacts. This occurs because the cached features from the previous scale lack the high-frequency refinements of the current scale, creating a semantic gap. In contrast, our strategy ensures that every token in the reconstructed map originates from the current processing stage, maintaining visual sharpness and coherence despite the slightly higher FID.

\begin{figure}[htbp]
    \centering
    \begin{minipage}[b]{0.55\textwidth}
        \centering
        \includegraphics[width=\textwidth]{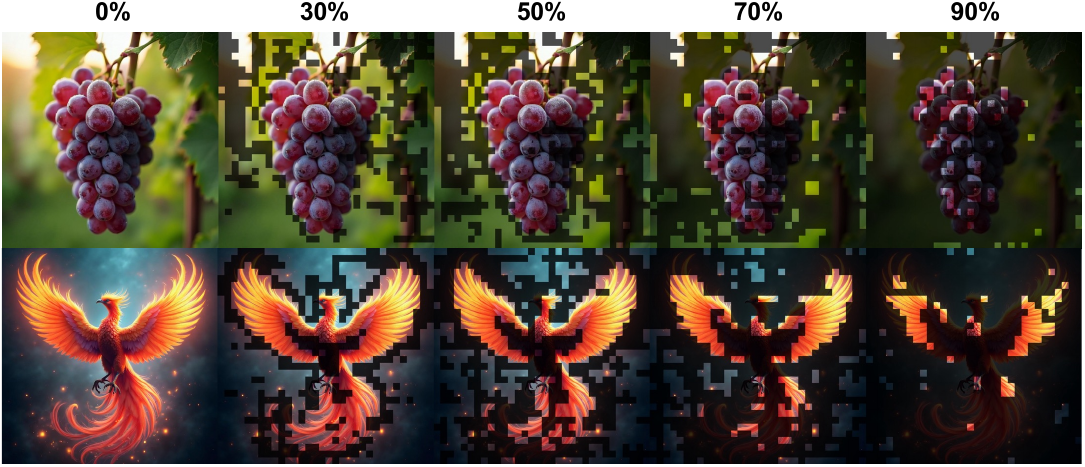}
        \caption{\textbf{Visualize prune masks under different ratios.}}
        \label{fig:mask_visualization}
    \end{minipage}
    \hfill
    \begin{minipage}[b]{0.42\textwidth}
        \centering
        \includegraphics[width=\textwidth]{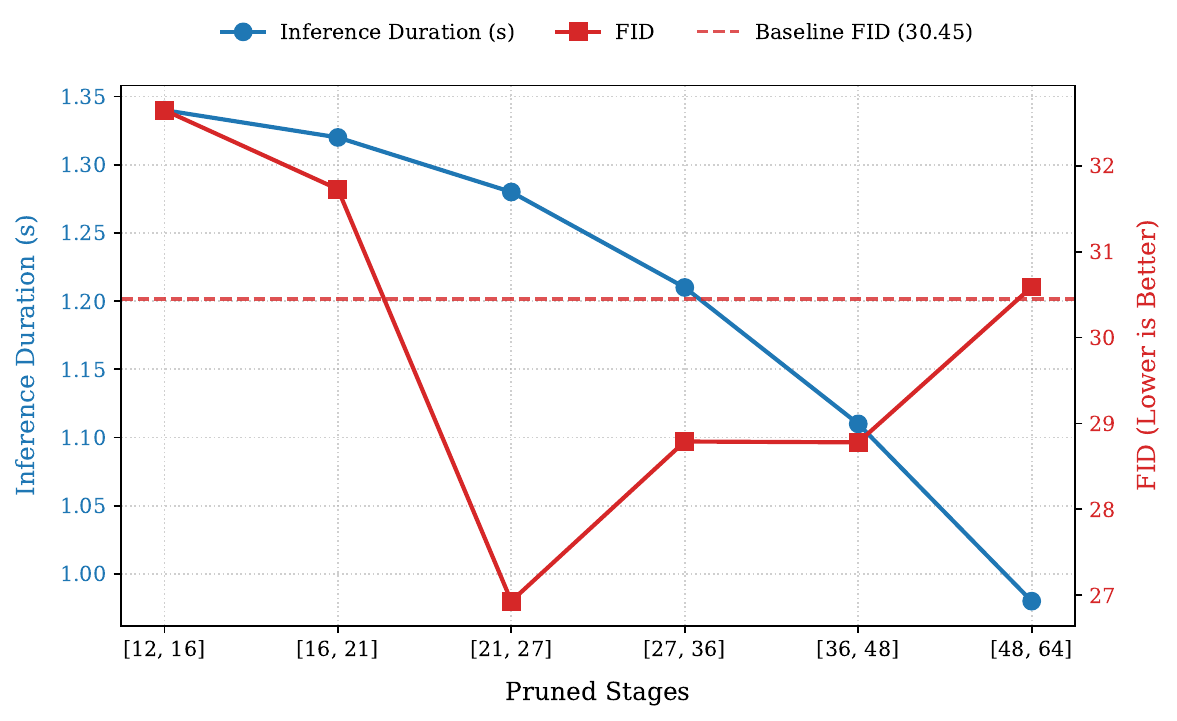}
        \caption{\textbf{Ablation on stage selection.}}
        \label{fig:ablate_stages}
    \end{minipage}
\end{figure}

\noindent \textbf{Sensitivity to Pruning Ratio.}
We evaluate the trade-off between efficiency and quality by varying the pruning ratio from 0 to 0.9 on the last two scales. Table~\ref{tab:ablation_prune_ratio} demonstrates that \methodNAME is highly robust; the generation quality remains stable even as the pruning ratio increases to 0.7, yielding a $1.43\times$ speedup. Performance only begins to degrade significantly at the extreme ratio of 0.9.
We further visualize the pruning masks in Figure~\ref{fig:mask_visualization}. The visualization reveals that our pruning strategy naturally acts as a saliency detector. At moderate ratios, the mask consistently retains tokens corresponding to the main subject and complex textures, while aggressively pruning the background. This selective retention allows the model to focus its computational budget on refining the salient object, explaining why high perceptual quality is maintained even when substantial portions of the background are discarded.

\noindent \textbf{Stage Selection Analysis.}
Finally, we validate our observation regarding the hierarchical importance of scales. We conducted experiments by pruning different pairs of adjacent scales with a fixed ratio of 0.7. Results are plotted in Figure~\ref{fig:ablate_stages}.
Pruning the early stages yields negligible speedups due to the small sequence length, yet causes a dramatic spike in FID. This confirms that early scales are structurally foundational and intolerant to error. Conversely, pruning the later stages results in significant latency reduction with minimal impact on generation quality. This supports our motivation to target the final, highly redundant stages for acceleration.

\section{Conclusion}

In this paper, we presented \methodNAME, a training-free token pruning framework designed to accelerate Visual Autoregressive models. To resolve the quadratic bottleneck in high-resolution generation, we introduced a dual-criteria selection strategy that integrates Principal Component Analysis with spatial high-pass filtering, ensuring the preservation of both global structural layout and local textural details. Additionally, we implemented a nearest-neighbor feature propagation mechanism to efficiently reconstruct dense feature maps without introducing semantic gaps. Extensive evaluations on text-to-image and text-to-video benchmarks demonstrate that \methodNAME achieves superior speed-quality trade-offs compared to state-of-the-art acceleration techniques, offering a robust solution for efficient generative inference.

\bibliographystyle{splncs04}
\bibliography{main}
\end{document}